\newcommand{\CLASSINPUTbaselinestretch}{1}
\newcommand{\CLASSINPUTbottomtextmargin}{0.8in}
\documentclass[journal]{IEEEtran}
\usepackage{etoolbox}
\makeatletter
\patchcmd{\maketitle}{\@copyrightspace}{}{}{}

\makeatother

\usepackage[]{graphicx}
\usepackage{stfloats}  
\usepackage{cite}  
\usepackage{psfrag}  
\usepackage{subfigure}
\usepackage{wrapfig}
\usepackage{epsfig}
\usepackage{url}
\usepackage{amsmath, amssymb}
\usepackage{array}
\usepackage{multicol}
\usepackage{algorithm}
\usepackage{algorithmic}
\usepackage{mathrsfs}
\usepackage{multirow}
\usepackage[normalem]{ulem}

\usepackage[utf8x]{inputenc} 
\usepackage{ucs} 
\usepackage{makeidx} 
\usepackage[dvipsnames,usenames]{color} 
\usepackage{array} 
\usepackage{colortbl} 
\usepackage{booktabs} 
\usepackage[justification=centering]{caption}

\definecolor{tableheading}{rgb}{0.9,0.9,0.9}
\definecolor{softblue}{rgb}{0.8,0.8,1} 
\arrayrulecolor{ForestGreen} 
\usepackage{tabu}
\usepackage{floatflt}

\usepackage{amsmath, amssymb}
\usepackage{tikz}

\usepackage{color,soul}

\begin{document}






\title{TiM-DNN: Ternary in-Memory accelerator for Deep Neural Networks\thanks{This work was supported in part by C-BRIC, one of six centers in JUMP, a Semiconductor Research Corporation (SRC) program sponsored by DARPA.}}
\author{\IEEEauthorblockN{Shubham Jain, Sumeet Kumar Gupta, Anand Raghunathan} \\
       \IEEEauthorblockA{School of Electrical and Computer Engineering, Purdue University \\ 
	          \{jain130,guptask,raghunathan\}@purdue.edu}
}

\sloppy

\maketitle

\begin{abstract}
The use of lower precision has emerged as a popular technique to optimize the compute and storage requirements of complex Deep Neural Networks (DNNs). In the quest for lower precision, recent studies have shown that ternary DNNs (which represent weights and activations by signed ternary values) represent a promising sweet spot, achieving accuracy close to full-precision networks on complex tasks. We propose TiM-DNN, a programmable in-memory accelerator that is specifically designed to execute ternary DNNs. TiM-DNN supports various ternary representations including unweighted \{-1,0,1\}, symmetric weighted \{-a,0,a\}, and asymmetric weighted \{-a,0,b\} ternary systems. The building blocks of TiM-DNN are {\em TiM tiles} --- specialized memory arrays that perform massively parallel signed ternary vector-matrix multiplications with a single access. TiM tiles are in turn composed of {\em Ternary Processing Cells} (TPCs), bit-cells that function as both ternary storage units and signed ternary multiplication units. We evaluate an implementation of TiM-DNN in 32nm technology using an architectural simulator calibrated with SPICE simulations and RTL synthesis. We evaluate TiM-DNN across a suite of state-of-the-art DNN benchmarks including both deep convolutional and recurrent neural networks. A 32-tile instance of TiM-DNN achieves a peak performance of 114 TOPs/s, consumes ~0.9W power, and occupies 1.96$mm^{2}$ chip area, representing a ~300X and ~388X improvement in TOPS/W and TOPS/$mm^{2}$, respectively, compared to an NVIDIA Tesla V100 GPU. In comparison to specialized DNN accelerators, TiM-DNN achieves 55X-240X and 160X-291X improvement in TOPS/W and TOPS/$mm^{2}$, respectively. Finally, when compared to a well-optimized near-memory accelerator for ternary DNNs, TiM-DNN demonstrates 3.9x-4.7x improvement in system-level energy and 3.2x-4.2x speedup, underscoring the potential of in-memory computing for ternary DNNs.
\end{abstract}

\section{Introduction}
\label{sec:introduction}
{\noindent}Deep Neural Networks (DNNs) have drastically advanced the field of machine learning by enabling super-human accuracies for many cognitive tasks involved in image, video, and natural language processing~\cite{wired-dnns}. However, the high computation and storage costs of DNNs severely limit their deployment in energy and cost-constrained devices~\cite{swagath-aspdac16}. 

The use of lower precision to represent the weights and activations in DNNs is a promising technique for improving the efficiency of DNN inference (evaluation of pre-trained DNN models)~\cite{AxNN,XNOR-Net,BC,DoReFa-Net,TC,TTQ,WRPN,TNN,HitNet,FGQ,QuantRNN,PACT}. Reduced bit-precision can lower all facets of energy consumption including computation, memory and data transfer. Current commercial hardware~\cite{nvidia-volta-v100,edge-tpu} already supports 8-bit and 4-bit fixed point formats, while recent research has continued the push towards even lower precision~\cite{XNOR-Net,BC,DoReFa-Net,TC,TTQ,TNN,WRPN,HitNet,FGQ}.

\begin{figure}[htb]
  \vspace*{-6pt}
  \centering
  \includegraphics[width=\columnwidth]{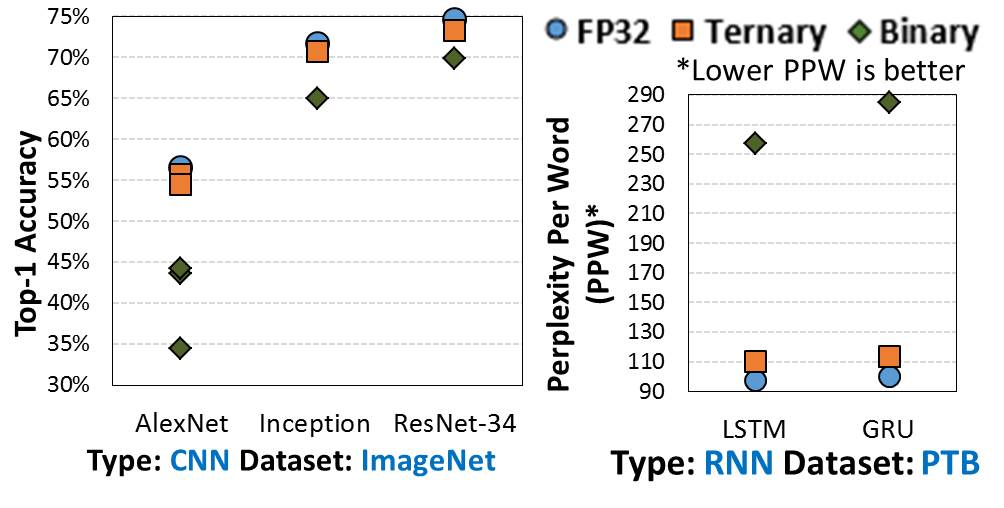}
  \vspace*{-8pt}
  \caption{Accuracy comparison of binary, ternary, and full-precision (FP32) DNNs~\cite{XNOR-Net,BC,TC,TTQ,WRPN,TNN,HitNet,FGQ,DoReFa-Net}}
  \label{fig:motivation}
  \vspace*{-6pt}
\end{figure}

Recent studies~\cite{XNOR-Net,BC,DoReFa-Net,TC,TTQ,WRPN,TNN,HitNet,FGQ,Compensated-DNN} suggest that ternary DNNs present a particularly attractive sweet-spot in the tradeoff between efficiency and accuracy. To illustrate this, Figure~\ref{fig:motivation} reports the accuracies of various state-of-the-art binary~\cite{XNOR-Net,BC,DoReFa-Net}, ternary~\cite{TC,TTQ,WRPN,TNN,HitNet,FGQ}, and full-precision (FP32) DNNs for image classification (ImageNet~\cite{ILSVRC15}) and language modeling (PTB~\cite{PTB}). We observe that the accuracy degradation of binary DNNs over the FP32 networks can be considerable (5-13\% for image classification, 150-180 PPW [Perplexity Per Word] for language modeling). In contrast, ternary DNNs achieve accuracy significantly better than binary networks, and result in minimal degradation (0.53\% for image classification) compared to FP32 networks. Motivated by these results, we focus on the design of a programmable accelerator for realizing state-of-the-art ternary DNNs. 

The multiply-and-accumulate (MAC) operation represents 95-99\% of total DNN computations. Consequently, the amount of energy and time spent on DNN computations can be drastically improved by using ternary processing elements (the energy of a MAC operation has a super-linear relationship with precision). However, when classical accelerator architectures ({\em e.g.}, TPUs and GPUs) are adopted to realize ternary DNNs, memory becomes the energy and performance bottleneck due to sequential (row-by-row) reads and leakage in un-accessed rows. In-memory computing~\cite{XNOR-RRAM,Binary-RRAM,reno,prime,spindle,puma,In-mem-classifier,Conv-RAM,RLui:2018,XNOR-SRAM,Xcel-RAM,sandwichRAM,ISSCC-2020-QLiu,ISSCC-2020-XSi,ISSCC-2020-Xue,ISSCC-2020-Qwong,ISSCC-2020-JWsu,ISSCC-2020-JYue,ISSCC-2019-Takemoto,ISSCC-2019-JWang,ISSCC-2019-XSi,ISSCC-2018-WKwa,ISSCC-2018-WChen,ISSCC-2018-TFWu,ISSCC-2018-Shanbhag} is an emerging computing paradigm that overcomes memory bottlenecks by integrating computations within the memory array itself, enabling greater parallelism and reducing the need to transfer data to/from memory. This work explores in-memory computing in the specific context of ternary DNNs and demonstrates that it leads to significant improvements in performance and energy efficiency.


While several efforts have explored in-memory accelerators in recent years, TiM-DNN differs in significant ways and is the first to apply in-memory computing (massively parallel vector-matrix multiplications within the memory array) to ternary DNNs using a new CMOS based bit-cell. Prior efforts have explored SRAM-based in-memory accelerators for binary networks~\cite{In-mem-classifier,Conv-RAM,RLui:2018,XNOR-SRAM,Xcel-RAM,sandwichRAM}. However, the restriction to binary networks is a significant limitation as binary networks known to date incur a large drop in accuracy as highlighted in Figure~\ref{fig:motivation}. Many in-memory accelerators use non-volatile memory (NVM) technologies such as PCM and ReRAM~\cite{XNOR-RRAM,Binary-RRAM,reno,prime,spindle,puma,mutliBitCiMISSCC2019} to realize in-memory dot product operations. While NVMs promise higher density and lower leakage than CMOS memories, they are still an emerging technology with open challenges such as large-scale manufacturing yield, limited endurance, high write energy, and errors due to device and circuit-level non-idealities~\cite{mnsim,rxnn}. Near-memory accelerators for ternary networks~\cite{TNN,BRein} have also been proposed, but their performance and energy are limited by sequential (row-by-row) memory access. SRAMs augmented with in-memory binary computation and additional near-memory logic have been proposed to perform higher precision computations in a bit-serial manner~\cite{neuralCache}. However, such an approach suffers from similar bottlenecks, limiting efficiency.

We propose TiM-DNN, a programmable in-memory accelerator that can realize massively parallel~\emph{signed ternary vector-matrix multiplications} per array access. TiM-DNN supports various ternary representations including unweighted \{-1,0,1\}, symmetric weighted \{-a,0,a\}, and asymmetric weighted \{-a,0,b\} systems, enabling it to execute a broad range of state-of-the-art ternary DNNs. This is motivated by recent efforts~\cite{TC,TTQ,WRPN,TNN,HitNet,FGQ} that show weighted ternary systems can achieve improved accuracies. The building block of TiM-DNN is a new memory cell called the Ternary Processing Cell (TPC), which functions as both a ternary storage unit and a scalar ternary multiplication unit. Using TPCs, we design TiM tiles, which are specialized memory arrays that execute signed ternary dot-product operations. TiM-DNN comprises of a plurality of TiM tiles arranged into banks, wherein all tiles compute signed vector-matrix multiplications in parallel.




We develop an architectural simulator for TiM-DNN, with array-level timing and energy models obtained from circuit-level simulations in 32nm CMOS technology. We evaluate TiM-DNN using a suite of 5 popular DNNs designed for image classification and language modeling tasks. A 32-tile instance of TiM-DNN achieves a peak performance of 114 TOPs/s, consumes ~0.9W power, and occupies 1.96$mm^{2}$ chip area, representing a ~300X improvement in TOPS/W compared to a state-of-the-art NVIDIA Tesla V100 GPU~\cite{nvidia-volta-v100}. In comparison to recent low-precision accelerators~\cite{neuralCache,BRein}, TiM-DNN achieves 55.2X-240X improvement in TOPS/W. Finally, TiM-DNN obtains 3.9x-4.7x improvement in system energy and 3.2x-4.2x improvement in performance over a highly-optimized near-memory accelerator for ternary DNNs. 

\vspace*{-0pt}

\vspace*{-0pt}
\section{Related Work }
\label{sec:relatedWork}
{\noindent}In recent years, several research efforts have focused on improving the energy efficiency and performance of DNNs at various levels of design abstraction. In this section, we limit our discussion to in-memory computing for DNNs~\cite{XNOR-RRAM,Binary-RRAM,reno,prime,spindle,puma,In-mem-classifier,Conv-RAM,RLui:2018,XNOR-SRAM,Xcel-RAM,neuralCache,dotSRAM,compute-mem,sttcim,neurocube}.    

\begin{table}[htb]
  \vspace*{-0pt}
  \centering
  \caption{Related work summary}
  \vspace*{-6pt}
  \includegraphics[width=\columnwidth]{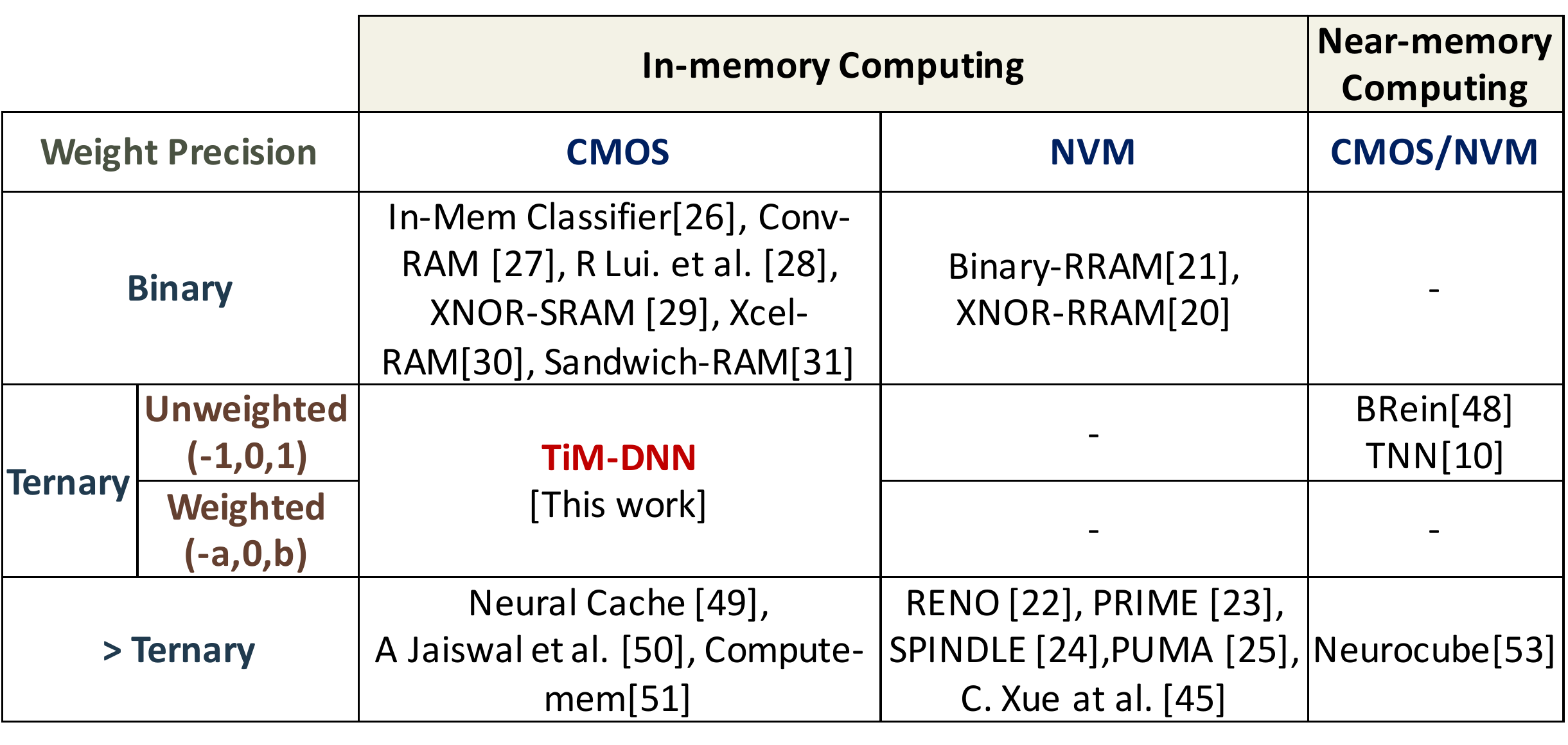}
  \label{tab:relatedWork}
  \vspace*{-6pt}
\end{table}

Table~\ref{tab:relatedWork} classifies prior in-memory computing efforts based on the memory technology [CMOS and Non-Volatile Memory (NVMs)] and the targeted precision (binary, ternary and super-ternary). Efforts on SRAM-based in-memory accelerators can be classified into those that target binary~\cite{In-mem-classifier,Conv-RAM,RLui:2018,XNOR-SRAM,Xcel-RAM,sandwichRAM} and high-precision (4-8 bits)~\cite{neuralCache,dotSRAM,compute-mem} DNNs. Accelerators targeting binary DNNs~\cite{In-mem-classifier,Conv-RAM,RLui:2018,XNOR-SRAM,Xcel-RAM} can execute massively parallel vector-matrix multiplication per array access. However, the restriction to binary networks is a significant limitation as binary networks known to date incur a large drop in accuracy as highlighted in Figure~\ref{fig:motivation}.
Efforts~\cite{neuralCache,dotSRAM,compute-mem} that target higher precision (4-8 bit) DNNs require multiple execution steps (array accesses) to realize signed dot-product operations, wherein both weights and activations are signed numbers. For example, Neural cache~\cite{neuralCache} computes bitwise Boolean operations in-memory but uses bit-serial near-memory arithmetic to realize multiplications, requiring several array accesses per multiplication operation (and many more to realize dot-products). Apart from in-memory computing efforts, Table~\ref{tab:relatedWork} also details efforts targeting near-memory accelerators for ternary networks~\cite{TNN,BRein}. However, the efficiency of these near-memory accelerators is limited by the on-chip memory, as they can enable only one memory row per access. Further, none of these efforts support asymmetric weighted \{-a,0,b\} ternary systems. Another group of efforts~\cite{XNOR-RRAM,Binary-RRAM,reno,prime,spindle,puma,mutliBitCiMISSCC2019} has focused on in-memory DNN accelerators using emerging Non-Volatile memory (NVM) technology such as PCM and ReRAM. Although NVMs promise density and low leakage relative to CMOS, they still face several open challenges such as large-scale manufacturing yield, limited endurance, high write energy, and errors due to device and circuit-level non-idealities~\cite{mnsim,rxnn}.

TiM-DNN is the first specialized and programmable in-memory accelerator for ternary DNNs that supports various ternary representations including unweighted \{-1,0,1\}, symmetric weighted \{-a,0,a\}, and asymmetric weighted \{-a,0,b\} ternary systems. TiM-DNN utilizes a new CMOS based bit-cell (i.e., TPC) and enables multiple memory rows simultaneously to realize massively parallel in-memory signed vector-matrix multiplications with ternary values per memory access, enabling efficient realization of ternary DNNs. As demonstrated in our experimental evaluation, TiM-DNN achieves significant speedup and energy improvements over near-memory ternary accelerators as well as higher-precision in-memory accelerators.




\vspace*{-0pt}
\section{TiM-DNN architecture}
\label{sec:timDNN}
{\noindent}In this section, we present the proposed TiM-DNN accelerator along with its building blocks,~\emph{i.e.}, Ternary Processing Cells and TiM tiles. 

\subsection{Ternary Processing Cell (TPC)}
\label{subsec:tpc}

\begin{figure}[htb]
  \vspace*{-6pt}
  \centering
  \includegraphics[width=\columnwidth]{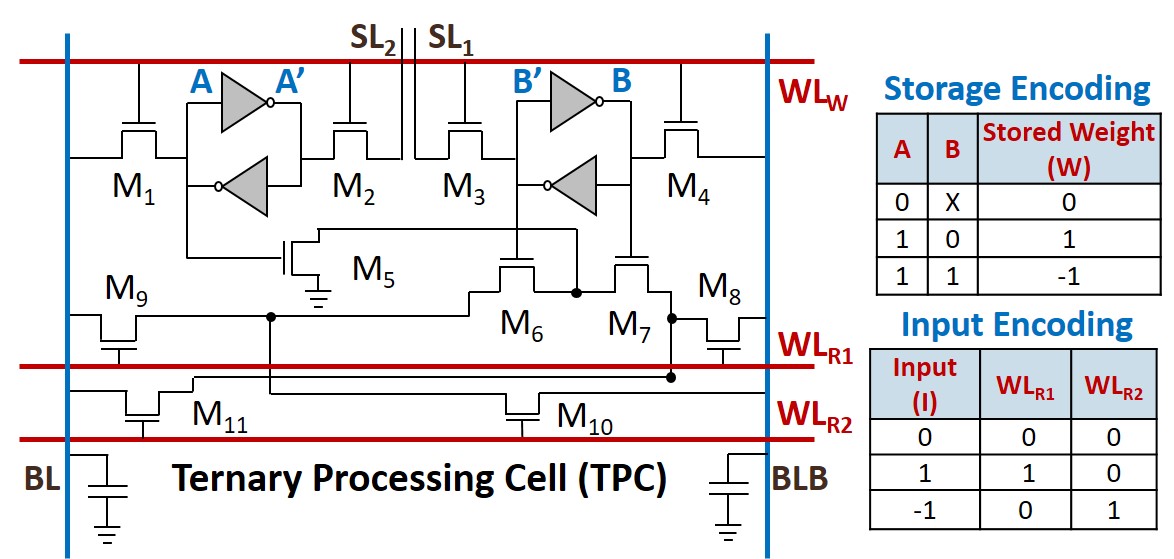}
  \vspace*{-10pt}
  \caption{Ternary Processing Cell (TPC) circuit, and weight and input encoding schemes}
  \label{fig:tpc}
  \vspace*{-6pt}
\end{figure}

To enable in-memory signed multiplication with ternary values, we present a new Ternary Processing Cell (TPC) that operates as both a ternary storage unit and a ternary scalar multiplication unit. Figure~\ref{fig:tpc} shows the proposed TPC circuit, which consists of two pairs of cross-coupled inverters for storing two bits (`A' and `B'), a write wordline ($WL_{W}$), two source lines ($SL_{1}$ and $SL_{2}$), two read wordlines ($WL_{R1}$ and $WL_{R2}$) and two bitlines (BL and BLB). A TPC supports two operations - write and scalar ternary multiplication. A write operation is performed by enabling $WL_{W}$ and driving the source-lines and the bitlines to either $V_{DD}$ or 0 depending on the data. We can write both bits simultaneously, with `A' written using BL and $SL_{2}$ and `B’ written using BLB and $SL_{1}$. Using both bits `A' and `B' a ternary value \{-1,0,1\} is inferred based on the storage encoding shown in Figure~\ref{fig:tpc} (table on the top right). For example, when A=0 the TPC stores W=0 regardless of the value of B. When A=1 and B=0 the TPC stores W=1, and when A=1 and B=1, W=-1.

\begin{figure}[htb]
  \vspace*{-6pt}
  \centering
  \includegraphics[width=0.9\columnwidth]{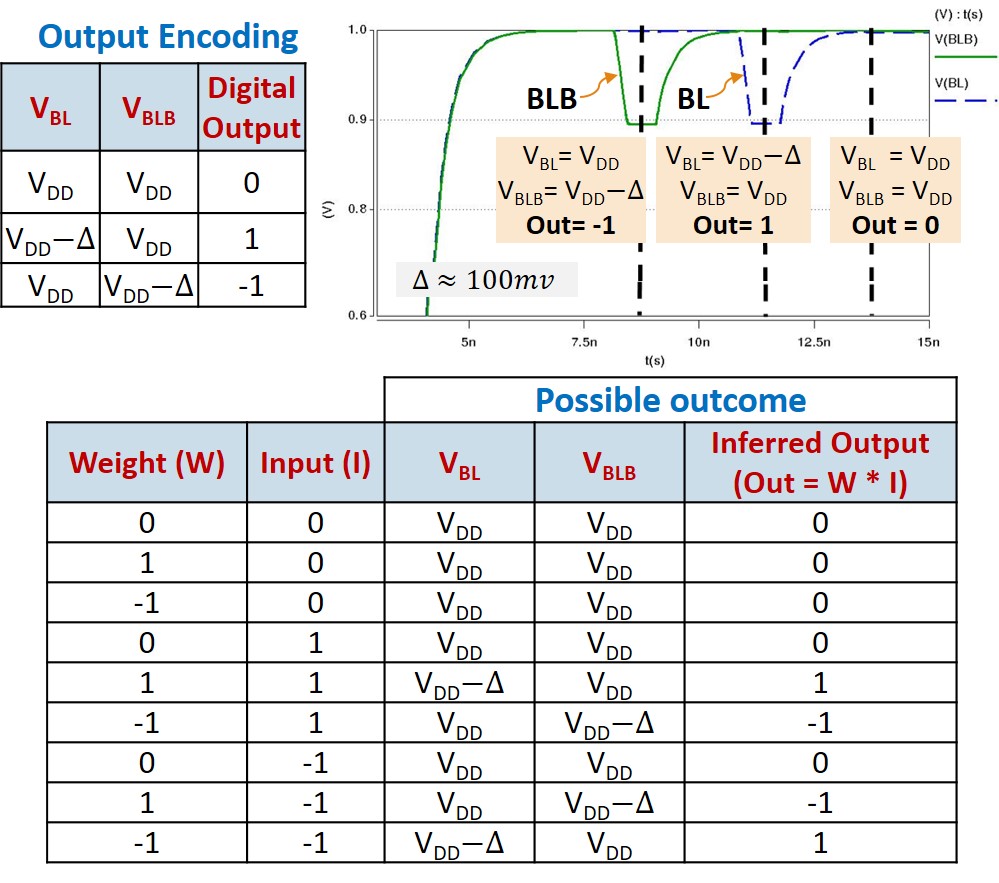}
  \vspace*{-6pt}
  \caption{Scalar multiplication using a TPC}
  \label{fig:scalarMul}
  \vspace*{-6pt}
\end{figure}

A scalar multiplication in a TPC is performed between a ternary input and the stored weight to obtain a ternary output. The bitlines are precharged to $V_{DD}$, and subsequently, the ternary inputs are applied to the read wordlines ($WL_{R1}$ and $WL_{R2}$) based on the input encoding scheme shown in Figure~\ref{fig:tpc} (table on the bottom right). The final bitline voltages ($V_{BL}$ and $V_{BLB}$) depend on both the input (I) and the stored weight (W). The table in Figure~\ref{fig:scalarMul} details the possible outcomes of the scalar ternary multiplication (W*I) with the final bitline voltages and the inferred ternary output (Out). 
For example, when W=0 or I=0, both bitlines remain at $V_{DD}$ and the output is inferred as 0 (W*I=0). When W=I=$\pm$1, BL discharges by a certain voltage, denoted by $\Delta$, and BLB remains at $V_{DD}$. This is inferred as Out=1. In contrast, when W=-I=$\pm$1, BLB discharges by $\Delta$ and BL remains at $V_{DD}$ producing Out=-1. The final bitline voltages are converted to a ternary output using single-ended sensing at BL and BLB. 
Figure~\ref{fig:scalarMul} depicts the output encoding scheme and the results of SPICE simulation of the scalar multiplication operation with various possible final bitline voltages. Note that the proposed TPC design uses separate read and write paths to avoid disturb failures during in-memory multiplications.   

\subsection{Dot-product computation using TPCs }
\label{subsec:dotProduct}

\begin{figure}[htb]
  \vspace*{-0pt}
  \centering
  \includegraphics[width=0.95\columnwidth]{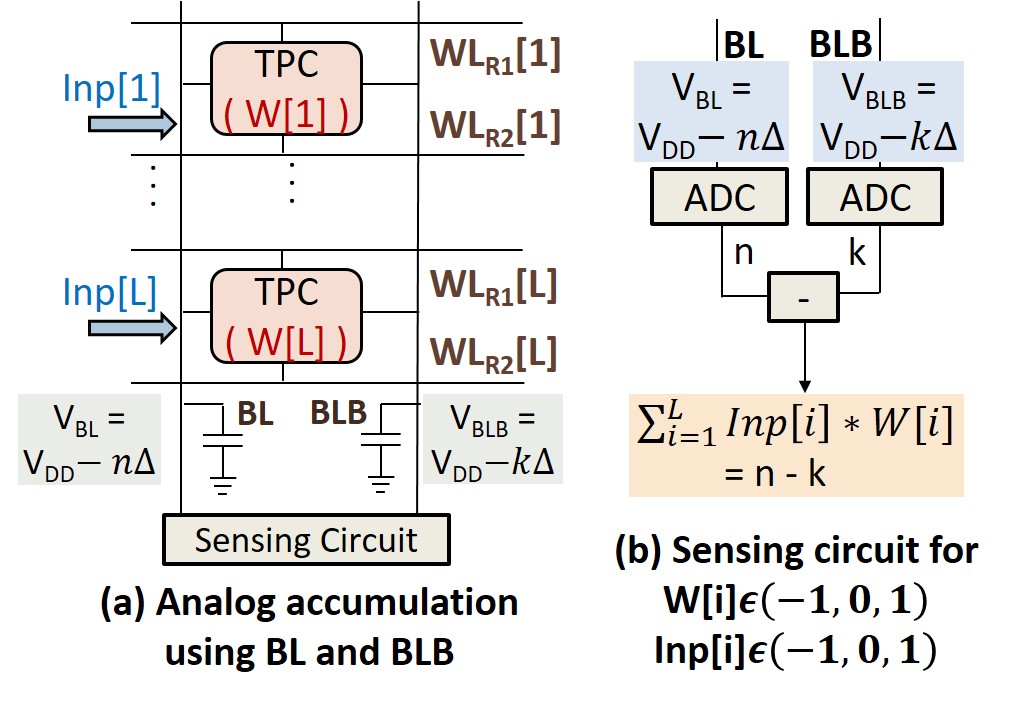}
  \vspace*{-6pt}
  \caption{Dot-product computation using TPCs: (a) Analog accumulation using BL and BLB, (b) sensing circuit for unweighted ternary encoding}
  \label{fig:dot-product}
  \vspace*{-6pt}
\end{figure}

Next, we extend the idea of realizing a scalar multiplication using the TPC to a dot-product computation. Figure~\ref{fig:dot-product}(a) illustrates the mapping of a dot-product operation ($\sum_{i=1}^{L} Inp[i]*W[i]$) to a column of TPCs with shared bitlines. The bitlines are first precharged to $V_{DD}$, and then the inputs (Inp) are applied to all TPCs simultaneously. The bitlines (BL and BLB) function as an analog accumulator, wherein the final bitline voltages ($V_{BL}$ and $V_{BLB}$) represent the sum of the individual TPC outputs. If $n$ out of $L$ TPCs produce output 1 and $k$ of $L$ TPCs produce output -1, the final bitline voltages will be roughly $V_{BL}=V_{DD}-n\Delta$ and $V_{BLB}=V_{DD}-k\Delta$. The bitline voltages are converted using Analog-to-Digital converters (ADCs) to yield digital values $n$ and $k$. For the unweighted ternary encoding where the weights and activations are encoded as \{-1,0,1\}, the final dot-product is $n-k$. Figure~\ref{fig:dot-product}(b) shows the sensing circuit required to realize dot-products with unweighted ternary encoding. Note that the node between M6 and M7 is floating (Figure~\ref{fig:tpc}). However, our simulations suggest that this floating node has no impact on the computed dot-products, as the bit-line capacitances are about three orders-of-magnitude larger than the M6-M7 node capacitance.

\begin{figure}[htb]
  \vspace*{-0pt}
  \centering
  \includegraphics[width=\columnwidth]{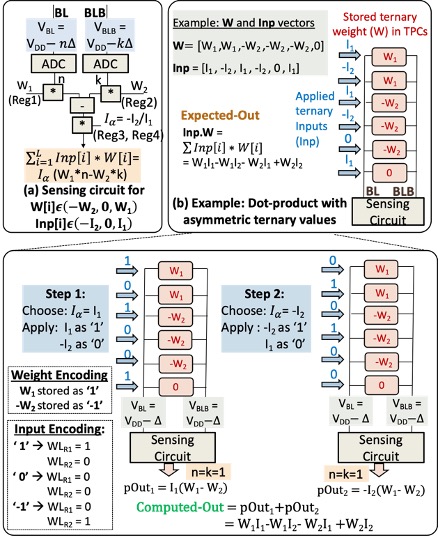}
  \vspace*{-10pt}
  \caption{(a) Sensing circuit for asymmetric weighted ternary system, (b) Example dot-product computation with asymmetric weighted ternary values }
  \label{fig:ExampleAsymmetric}
  \vspace*{-6pt}
\end{figure}

We can also realize dot-products with a more general ternary encoding represented by asymmetric weighted \{-a,0,b\} values. Support for a more general ternary encoding is motivated by recent efforts~\cite{TTQ} showing that weighted ternary systems can achieve improved accuracies. Figure~\ref{fig:ExampleAsymmetric}(a) shows the sensing circuit that enables dot-product with asymmetric ternary weights \{$-W_{2},0,W_{1}$\} and inputs \{$-I_{2},0,I_{1}$\}. As shown, the ADC outputs are scaled by the corresponding weights ($W_{1}$ and $W_{2}$), and subsequently, an input scaling factor ($I_{\alpha}$) is applied to yield `$I_{\alpha}$($W_{1}$*n-$W_{2}$*k)'. In contrast to dot-products with unweighted values, we require two execution steps to realize dot-products with the asymmetric ternary system, wherein each step computes a partial dot-product (pOut). Figure~\ref{fig:ExampleAsymmetric}(b) details these two steps using an example. In step 1, we choose $I_{\alpha}$=$I_{1}$, and apply $I_{1}$ and $I_{2}$ as `1' and `0', respectively, resulting in a partial output (pOut) given by $pOut_{1}$ = $I_{1}$($W_{1}$*n-$W_{2}$*k). In step 2, we choose $I_{\alpha}$=-$I_{2}$, and apply $I_{1}$ and $I_{2}$ as `0' and `1', respectively, to yield $pOut_{2}$ = -$I_{2}$($W_{1}$*n-$W_{2}$*k). The final dot-product is given by `$pOut_{1}$+$pOut_{2}$'.


\begin{figure}[htb]
  \vspace*{-4pt}
  \centering
  \includegraphics[width=\columnwidth]{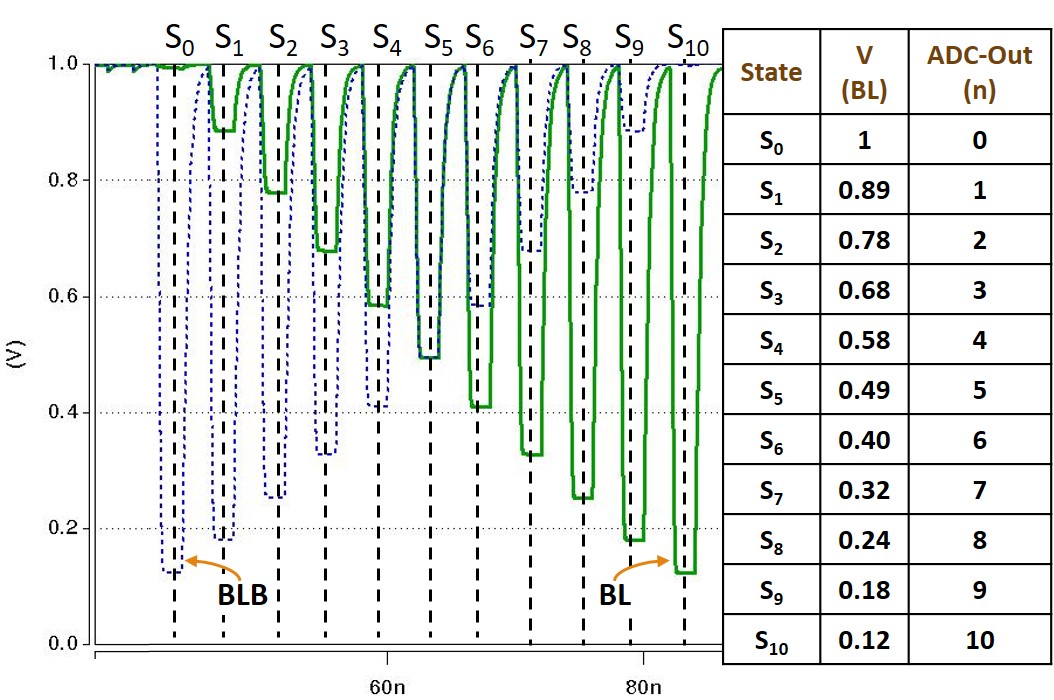}
  \vspace*{-4pt}
  \caption{Dot-product circuit simulation}
  \label{fig:cirSim}
  \vspace*{-6pt}
\end{figure}
 
\begin{figure*}[htb]
  \vspace*{-8pt}
  \centering
  \includegraphics[width=\textwidth]{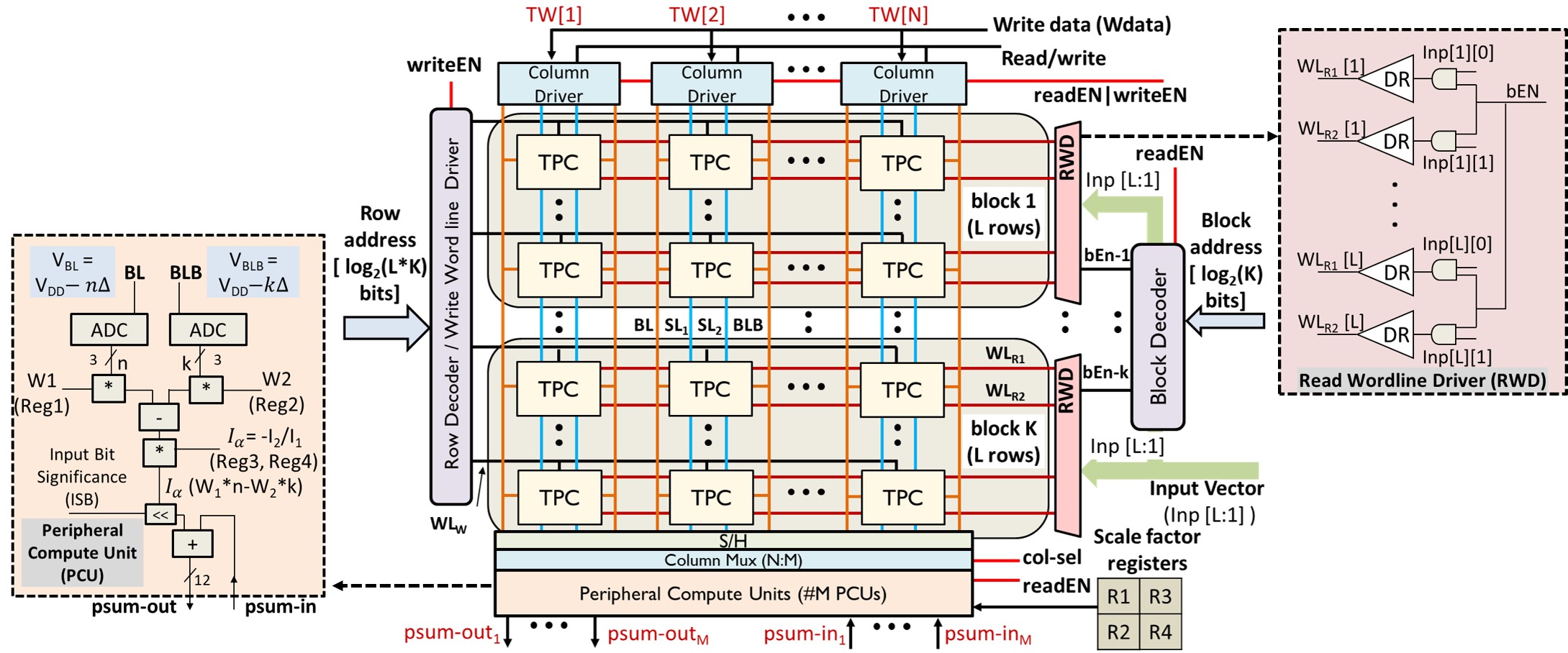}
  \vspace*{-8pt}
  \caption{Ternary in-Memory processing tile}
  \label{fig:timTile}
  \vspace*{-8pt}
\end{figure*}

To validate the dot-product operation, we perform a detailed SPICE simulation to determine the possible final voltages at BL ($V_{BL}$) and BLB ($V_{BLB}$). Figure~\ref{fig:cirSim} shows various BL states ($S_{0}$ to $S_{10}$) and the corresponding value of $V_{BL}$ and $n$. Note that the possible values for $V_{BLB}$ (which represents $k$) and $V_{BL}$ (which represents $n$) are identical, as BL and BLB are symmetric. The state $S_{i}$ refers to the scenario where $i$ out of $L$ TPCs compute an output of `1'. We observe that from $S_{0}$ to $S_{7}$ the average sensing margin ($\Delta$) is 96mv. The sensing margin decreases to 60-80mv for states $S_{8}$ to $S_{10}$, and beyond $S_{10}$ the bitline voltage ($V_{BL}$) saturates. Therefore, we can achieve a maximum of 11 BL states ($S_{0}$ to $S_{10}$) with sufficiently large sensing margin required for sensing reliably under process variations~\cite{XNOR-SRAM}. The maximum value of $n$ and $k$ is thus 10, which in turn determines the number of TPCs ($L$) that can be enabled simultaneously. Setting $L = n_{max} = k_{max}$ would be a conservative choice. However, exploiting the weight and input sparsity of ternary DNNs~\cite{WRPN,HitNet,FGQ}, wherein 40\% or more of the weights and inputs are zeros, and the fact that non-zero outputs are distributed between `1' and `-1', we choose a design with $n_{max}$ = 8, and $L$ = 16. Our experiments indicate that this choice has no impact on DNN accuracy compared to the conservative case. We also evaluate the impact of process variations on the dot-product operations realized using TPCs, and provide experimental results in Section~\ref{subsec:varAnalysis}. 

\subsection{TiM tile}
\label{subsec:timTile}

We now present the TiM tile, a specialized memory array designed using TPCs to realize massively parallel vector-matrix multiplications with ternary values. Figure~\ref{fig:timTile} details the tile design, which consists of a 2D array of TPCs, a row decoder and write wordline driver (for write operations), a block decoder (for compute operations), Read Wordline Drivers (RWDs), column drivers, a sample and hold (S/H) unit, a column mux, Peripheral Compute Units (PCUs), and scale factor registers. The TPC array contains $L*K*N$ TPCs, arranged in $K$ blocks and $N$ columns, where each block contains $L$ rows. As shown in the Figure, TPCs in the same row (column) share wordlines (bitlines and source-lines). The tile supports two major functions, (i) row-by-row write operations, and (ii) vector-matrix multiplication. A write operation is performed by activating a write wordline ($WL_{W}$) using the row decoder and driving the bitlines and source-lines. During a write operation, $N$ ternary words (TWs) stored in one row of the array are written in parallel. In contrast to the row-wise write operation, a vector-matrix multiplication operation is realized at the block granularity, wherein $N$ dot-product operations each of vector length $L$ are executed in parallel. The block decoder selects a block for the vector-matrix multiplication, and RWDs apply the ternary inputs. During the vector-matrix multiplication, TPCs in the same row share a ternary input (Inp), and TPCs in the same column produce partial sums for the same output. As discussed in section~\ref{subsec:dotProduct}, accumulation is performed in the analog domain using the bitlines (BL and BLB). In one access, a TiM tile can compute the vector-matrix product $\bf{Inp.W}$, where {\bf Inp} is a vector of length $L$ and {\bf W} is a matrix of dimension $L \times N$ stored in TPCs. The accumulated outputs at each column are stored using a sample and hold (S/H) unit and get digitized using PCUs. To attain higher area efficiency, we utilize $M$ PCUs per tile ($M < N$) by matching the bandwidth of the PCUs to the bandwidth of the TPC array and operating the PCUs and TPC array as a two-stage pipeline. Next, we discuss the TiM tile peripherals in detail. 
   
{\bf \noindent Read Wordline Driver (RWD).} Figure~\ref{fig:timTile} shows the RWD logic that takes a ternary vector ({\bf Inp}) and block enable (bEN) signal as inputs and drives all `L' read wordlines ($WL_{R1}$ and $WL_{R2}$) of a block. The block decoder generates the bEN signal based on the block address that is an input to the TiM tile. $WL_{R1}$ and $WL_{R2}$ are activated using the input encoding scheme shown in Figure~\ref{fig:tpc} (Table on the bottom).


{\bf \noindent Peripheral Compute Unit (PCU).} Figure~\ref{fig:timTile} shows the logic for a PCU, which consists of two ADCs and a few small arithmetic units (adders and multipliers). The primary function of PCUs is to convert the bitline voltages to digital values using ADCs. However, PCUs also enable other key functions such as partial sum reduction, and weight (input) scaling for weighted ternary encoding (-$W_{2}$,0,$W_{1}$) and (-$I_{2}$,0,$I_{1}$). Although the PCU can be simplified if $W_{2}$=$W_{1}$=1 or/and  $I_{2}$=$I_{1}$=1, in this work, we target a programmable TiM tile that can support various state-of-the-art ternary DNNs. To further generalize, we use a shifter to support DNNs with ternary weights and higher precision activations~\cite{WRPN,FGQ}. The activations are evaluated bit-serially using multiple TiM accesses. Each access uses an input bit, and we shift the computed partial sum based on the input bit significance using the shifter. TiM tiles have scale factor registers (shown in Figure~\ref{fig:timTile}) to store the weight and the activation scale factors that vary across layers within a network.

\subsection{TiM-DNN accelerator architecture}
\label{subsec:accelerator}

\begin{figure}[htb]
  \vspace*{-6pt}
  \centering
  \includegraphics[width=\columnwidth]{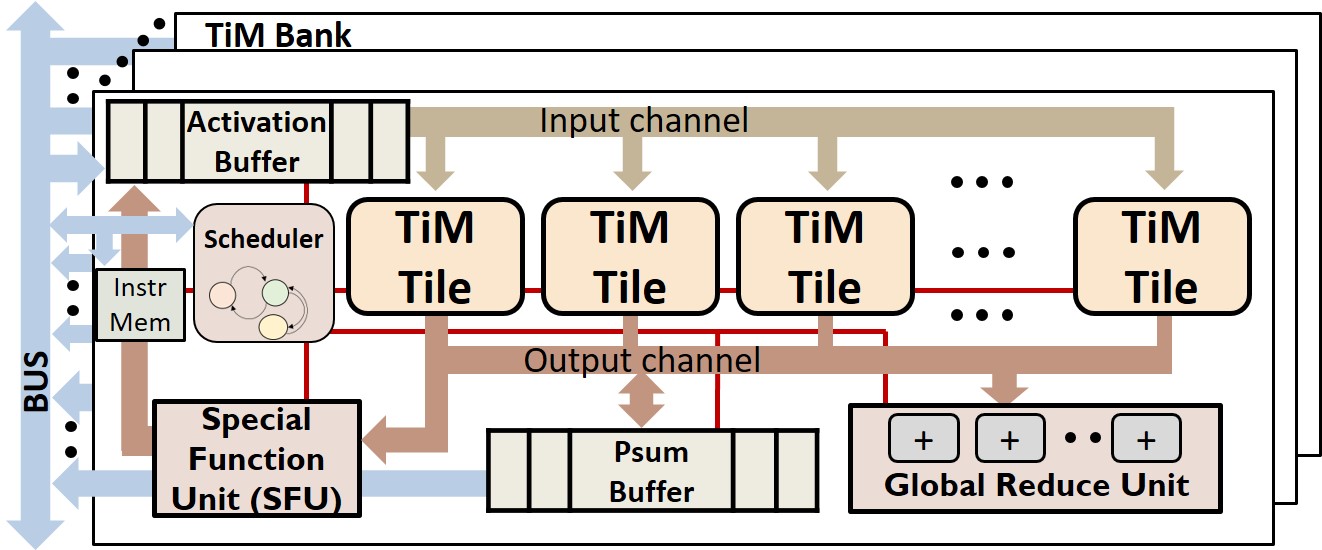}
  \vspace*{-6pt}
  \caption{TiM-DNN accelerator architecture}
  \label{fig:TiMaccelerator}
  \vspace*{-0pt}
\end{figure}

{\noindent}Figure~\ref{fig:TiMaccelerator} shows the proposed TiM-DNN accelerator, which has a hierarchical organization with multiple banks, wherein each bank is comprised of several TiM tiles, an activation buffer, a partial sum (Psum) buffer, a global Reduce Unit (RU), a Special Function Unit (SFU), an instruction memory (Inst Mem), and a Scheduler. The compute time and energy in Ternary DNNs are heavily dominated by vector-matrix multiplications, which are realized using TiM tiles. Other DNN computations, ~\emph{viz.}, ReLU, pooling, normalization, Tanh and Sigmoid are performed by the SFU. The partial sums produced by different TiM tiles are reduced using the RU, whereas the partial sums produced by separate blocks within a tile are reduced using PCUs (as discussed in section~\ref{subsec:timTile}). TiM-DNN has a small instruction memory and a scheduler that reads instructions and orchestrates operations inside a bank. TiM-DNN also contains activation and Psum buffers to store activations and partial sums, respectively.   
  
\begin{figure}[htb]
  \vspace*{-6pt}
  \centering
  \includegraphics[width=0.95\columnwidth]{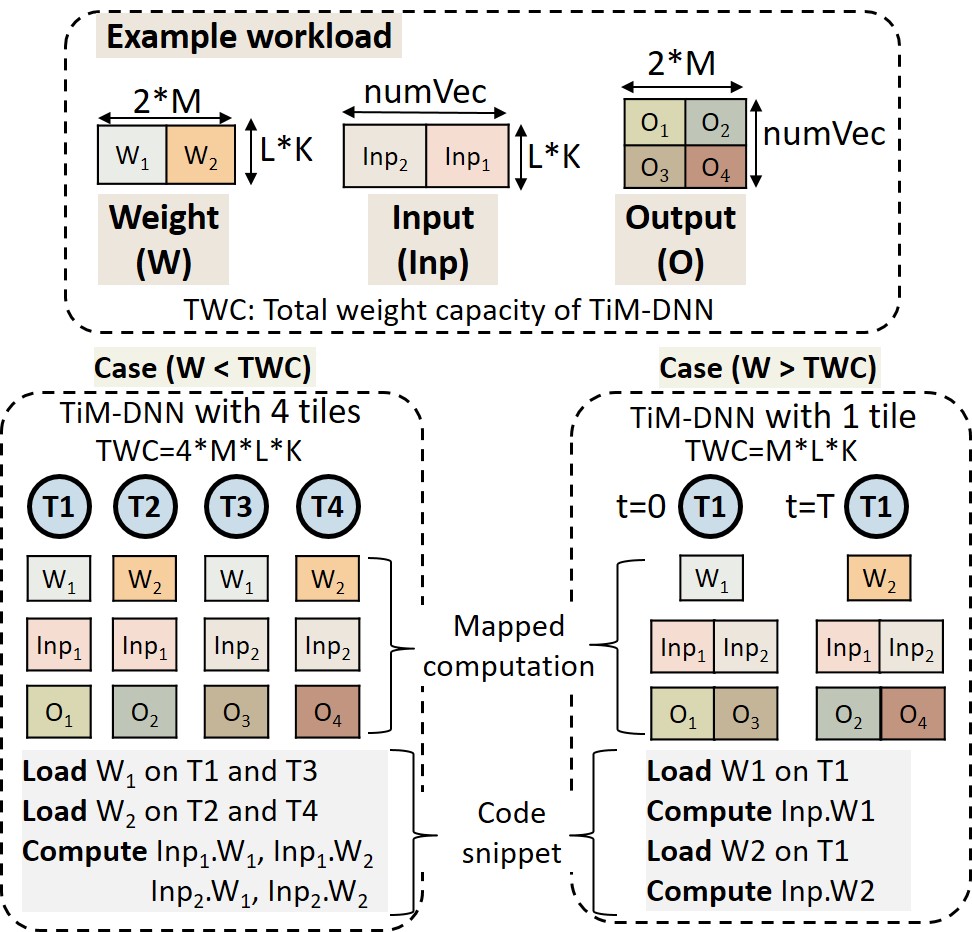}
  \vspace*{-0pt}
  \caption{TiM-DNN mapping: Example}
  \label{fig:mapping}
  \vspace*{-0pt}
\end{figure}

{\bf \noindent Mapping.} DNNs can be mapped to TiM-DNN sparially or temporally. The networks that fit on TiM-DNN entirely are mapped spatially, wherein the weight matrix of each convolution (Conv) and fully-connected (FC) layer is partitioned and mapped to (one or more) dedicated TiM tiles, and the network executes in a layer-wise pipelined fashion. In contrast, networks that cannot fit on TiM-DNN at once are executed using the temporal mapping strategy, wherein we execute Conv and FC layers sequentially using all TiM tiles. The weight matrix (W) of each CONV/FC layer could be either smaller or larger than the total weight capacity (TWC) of TiM-DNN. Figure~\ref{fig:mapping} illustrates the two scenarios using an example workload (vector-matrix multiplication) that is executed on two separate TiM-DNN instances differing in the number of TiM tiles. As shown, when ($W \leq$ TWC) the weight matrix partitions ($W_{1}$ \& $W_{2}$) are replicated and loaded to multiple tiles, and each TiM tile computes on input vectors in parallel. In contrast, when ($W >$ TWC), the operations are executed sequentially using multiple steps. In our evaluations, we mapped the CNN benchmarks using the temporal mapping strategy as they do not fit on TiM-DNN at once. In contrast, our RNN benchmarks fit on TiM-DNN entirely and are hence mapped spatially.




\vspace*{-0pt}
\section{Experimental Methodology}
\label{sec:exptsetup}
{\noindent}In this section, we present the experimental methodology used to evaluate TiM-DNN.  

\begin{figure}[htb]
  \vspace*{-0pt}
  \centering
  \includegraphics[width=\columnwidth]{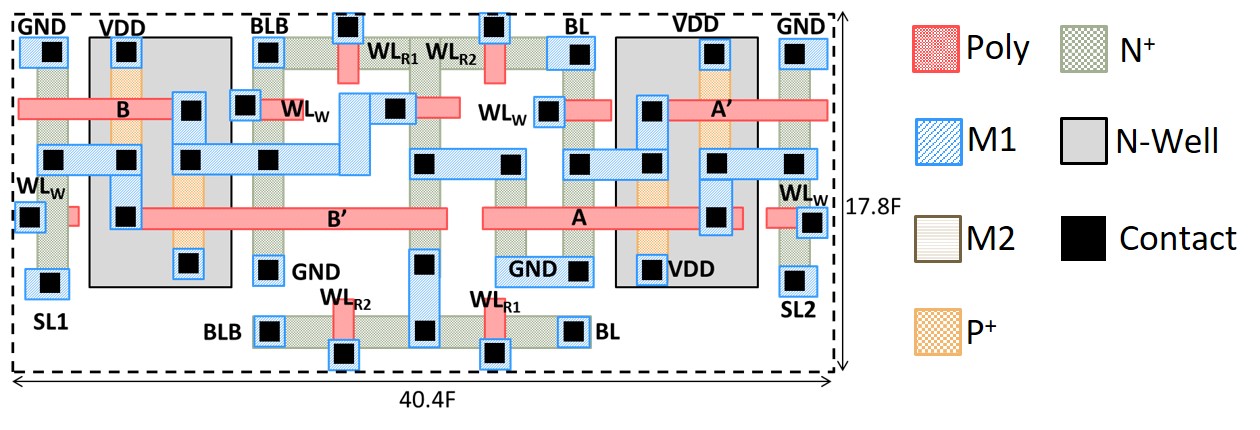}
  \vspace*{-6pt}
  \caption{Ternary Processing Cell (TPC) layout}
  \label{fig:layout1}
  \vspace*{-0pt}
\end{figure}

{\bf\noindent TiM tile modeling.} We performed detailed SPICE simulations to estimate the tile-level energy and latency for the write and vector-matrix multiplication operations. The simulations were performed using 32nm bulk CMOS technology and PTM models. We used 3-bit flash ADCs to convert bitline voltages to digital values. To estimate the area and latency of digital logic both within the tiles (PCUs and decoders) and outside the tiles (SFU and RU), we synthesized RTL implementations using Synopsys Design Compiler and estimated power consumption using Synopsys Power Compiler. We performed a layout of the TPC (Figure~\ref{fig:layout1}) to estimate its area, which is about $720F^{2}$ (where F is the minimum feature size). We also performed variation analysis to estimate error rates due to incorrect sensing by considering variations in transistor $V_{T}$ ($\sigma$/$\mu$=5\%)~\cite{kuhn2011}.


 \vspace*{+4pt}
 
{\bf\noindent System-level simulation.} We developed an architectural simulator to estimate application-level energy and performance benefits of TiM-DNN. The simulator maps various DNN operations,~\emph{viz.}, vector-matrix multiplications, pooling, Relu, etc. to TiM-DNN components and produces execution traces consisting of off-chip accesses, write and vector-matrix multiply operations in TiM tiles, buffer reads and writes, and RU and SFU operations. Using these traces and the timing and energy models from circuit simulation and synthesis, the simulator computes the application-level energy and performance.

{\bf\noindent TiM-DNN parameters.} Table~\ref{tab:microParams} details the micro-architectural parameters for the instance of TiM-DNN used in our evaluation, which contains 32 TiM tiles, with each tile having 256x256 TPCs. The SFU consists of 64 Relu units, 8 vector processing elements (vPE) with 4 lanes each, 20 special function processing elements (SPEs), and 32 Quantization Units (QU). SPEs compute special functions such as Tanh and Sigmoid. The output activations are quantized to ternary values using QUs. The latency of the dot-product operation is 2.3 ns. TiM-DNN can achieve a peak performance of 114 TOPs/sec, consumes $\sim$0.9 W power, and occupies $\sim$1.96 $mm^{2}$ chip area.

\begin{table}[htb]
  \vspace*{-0pt}
  \centering
  \caption{TiM-DNN micro-architectural parameters}
  \vspace*{-0pt}
  \includegraphics[width=\columnwidth]{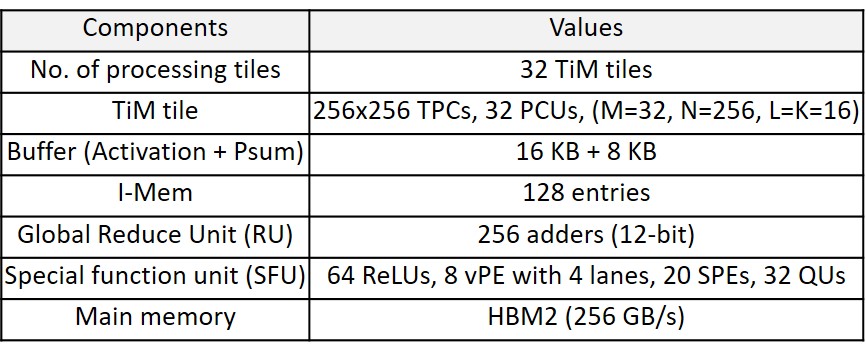}
  \label{tab:microParams}
  \vspace*{-0pt}
\end{table}

 \vspace*{+4pt}
 
  \vspace*{-2pt}
  \begin{floatingfigure}[r]{0.5\columnwidth}
  \centering
  \includegraphics[width=0.5\columnwidth]{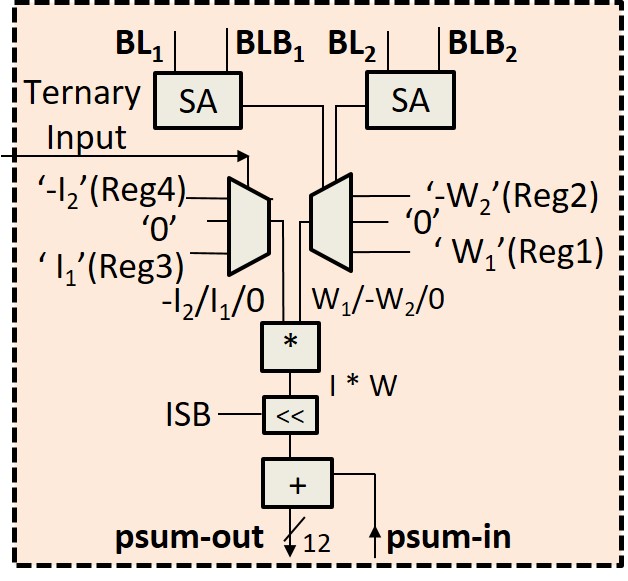}
  \vspace*{-6pt}
  \caption{Near-memory compute unit for the baseline design}
  \label{fig:PCUSRAMtile}
  \vspace*{-6pt}
\end{floatingfigure}

{\bf\noindent Baseline.} The processing efficiency (TOPS/W) of TiM-DNN is ~300X better than NVIDIA's state-of-the-art Volta V100 GPU~\cite{nvidia-volta-v100}. The large improvement is to be expected, since the GPU is not specialized for ternary DNNs. In comparison to a recently proposed near-memory ternary accelerator~\cite{BRein}, TiM-DNN achieves 55.2X improvement in TOPS/W. To isolate the benefits exclusively due to in-memory computations, we design a well-optimized near-memory ternary DNN accelerator. This baseline accelerator differs from TiM-DNN in only one aspect --- tiles consist of  regular SRAM arrays (256x512) with 6T bit-cells and near-memory compute (NMC) units (shown in Figure~\ref{fig:PCUSRAMtile}), instead of the TiM tiles. Note that, to store a ternary word using the SRAM array, we require two 6T bit-cells. The baseline tiles are smaller than TiM tiles by 0.52x, therefore, we use two baselines designs. (i) An iso-area baseline with 60 baseline tiles and the overall accelerator area is same as TiM-DNN.(ii) An iso-capacity baseline with the same weight storage capacity (2 Mega ternary words) as TiM-DNN. We note that the baseline is well-optimized, and our iso-area baseline can achieve 21.9 TOPs/sec, reflecting an improvement of 17.6X in TOPs/sec over the near-memory accelerator for ternary DNNs proposed in~\cite{BRein}. 


\vspace*{+4pt}

{\bf\noindent DNN Benchmarks.} We evaluate the system-level energy and performance benefits of TiM-DNN using a suite of DNN benchmarks listed in Table~\ref{tab:benchmarks}. We use state-of-the-art convolutional neural networks (CNNs), viz., AlexNet, ResNet-34, and Inception to perform image classification on ImageNet. We also evaluate recurrent neural networks (RNN), specifically an LSTM and a GRU that perform language modeling task on the Penn Tree Bank (PTB) dataset~\cite{PTB}. Table~\ref{tab:benchmarks} also details the activation precision and accuracy of these ternary networks.

\begin{table}[htb]
  \vspace*{-0pt}
  \centering
  \caption{DNN benchmarks}
  \vspace*{-6pt}
  \includegraphics[width=\columnwidth]{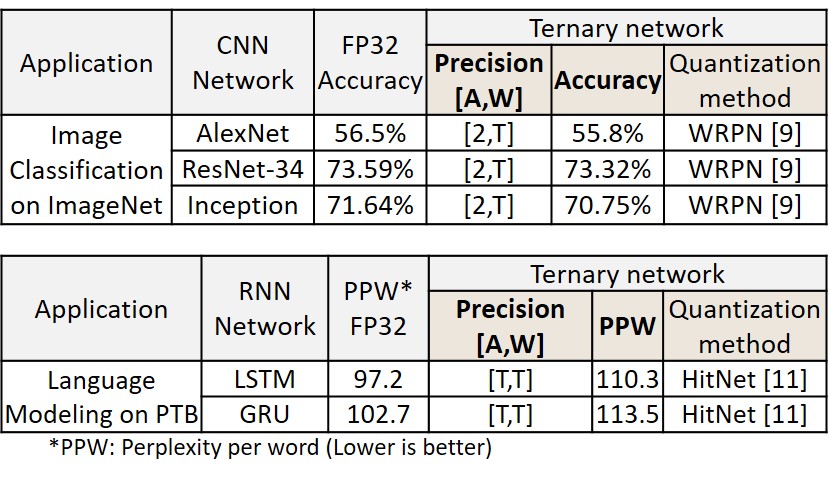}
  \label{tab:benchmarks}
  \vspace*{-6pt}
\end{table}

\vspace*{-0pt}
\section{Results}
\label{sec:results}
In this section, we present various results that quantify the improvements obtained by TiM-DNN. We also compare TiM-DNN with other state-of-the-art DNN accelerators.


\begin{table}[htb]
  \vspace*{-0pt}
  \centering
  \caption{Comparison of TiM-DNN with other DNN accelerators}
  \vspace*{-0pt}
  \includegraphics[width=\columnwidth]{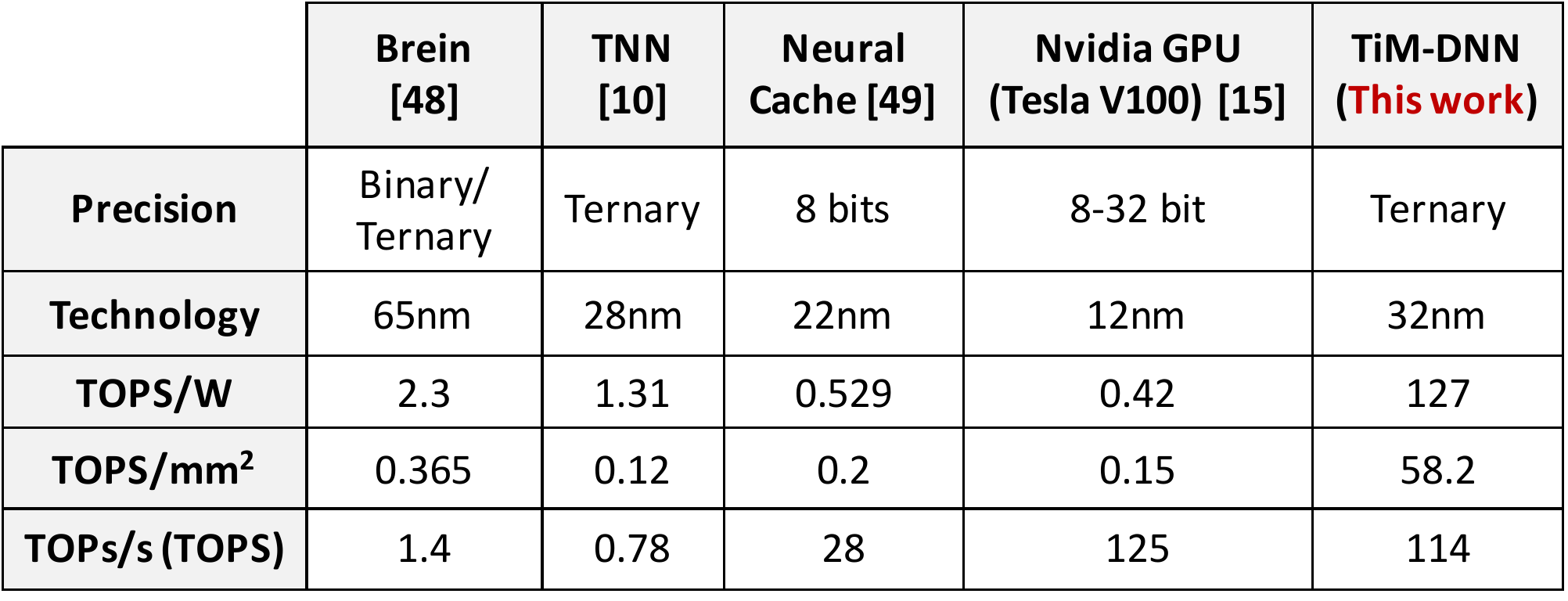}
  \label{tab:CompRelatedWork}
  \vspace*{-0pt}
\end{table}

\subsection{Comparison with prior designs:}
\label{subsec:GPUcomparision}
  
{\bf \noindent System-level.} We first quantify the advantages of TiM-DNN over prior DNN accelerators using processing efficiencies (TOPS/W and TOPS/$mm^{2}$) as our metric. Table~\ref{tab:CompRelatedWork} details 4 prior DNN accelerators including - (i) Nvidia Tesla V100~\cite{nvidia-volta-v100}, a state-of-the-art GPU, (ii) Neural-Cache~\cite{neuralCache}, a design using bitwise in-memory operations and bit-serial near-memory arithmetic to realize dot-products, as well as (iii) BRein~\cite{BRein} and (iv) TNN~\cite{TNN}, which are recently proposed near-memory accelerators for ternary DNNs. As shown, TiM-DNN achieves substantial improvements in both TOPS/W and TOPS/$mm^{2}$ over all baselines. One reason for the improvement is that unlike some of the baselines, TiM-DNN is specialized to execute ternary networks. Moreover, GPUs, near-memory accelerators~\cite{BRein}, and Neural-Cache~\cite{neuralCache} are still limited by memory bandwidth, since they can access one~\cite{nvidia-volta-v100,BRein} or at most two~\cite{neuralCache} memory rows from each array. In contrast, TiM-DNN offers additional parallelism by simultaneously accessing $L=16$  memory rows to compute in-memory vector-matrix multiplications.

\begin{table}[htb]
  \vspace*{-0pt}
  \centering
  \caption{Array-level comparision: TiM processing tile with prior designs}
  \vspace*{-0pt}
  \includegraphics[width=\columnwidth]{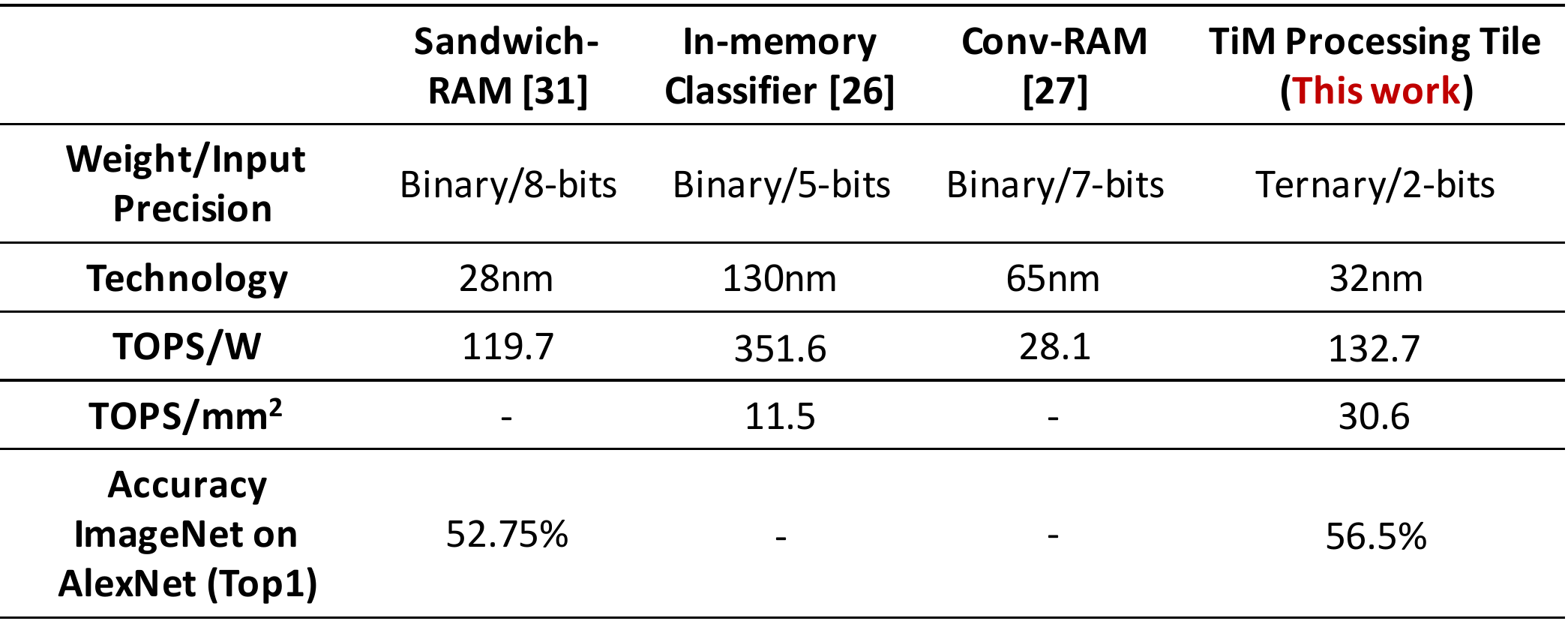}
  \label{tab:arrayLevelComp}
  \vspace*{-0pt}
\end{table}

{\bf \noindent Array-level.} Next, we compare the TiM processing tile with prior array-level designs~\cite{In-mem-classifier,Conv-RAM,sandwichRAM} for performing in-memory dot-product computations. Table~\ref{tab:arrayLevelComp} shows the comparison using TOPS/W, TOPS/$mm^{2}$, and ImageNet accuracy on AlexNet. Although some of these designs can achieve higher processing efficiency, their restriction to binary weights is a significant limitation as binary networks incur a large drop in accuracy (5-13\% for image classification) as highlighted in Figure~\ref{fig:motivation}. In contrast, ternary DNNs achieve accuracy significantly better than binary DNNs and much closer (0.53\% for image classification) to full-precision (FP32) networks. 

\subsection{Benefits of in-memory computing}
\label{subsec:systemPeformance}
The efficiency of TiM-DNN arises from being specialized to ternary processing, as well from in-memory computing. In order to specifically assess the benefits from in-memory computing, we compare the performance and energy of TiM-DNN to two well-optimized near-memory accelerator designs. These baselines are derived from TiM-DNN by replacing the TiM tiles with near-memory processing tiles, and consist of two variants, {\em viz.} Iso-capacity and Iso-area designs.

\begin{figure}[htb]
  \vspace*{-0pt}
  \centering
  \includegraphics[width=\columnwidth]{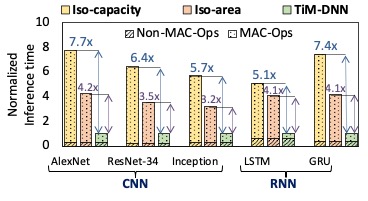}
  \vspace*{-6pt}
  \caption{Performance benefits of TiM-DNN}
  \label{fig:perfBenefits}
  \vspace*{-0pt}
\end{figure}

Figure~\ref{fig:perfBenefits} shows the two major components of the normalized inference time which are MAC-Ops (vector-matrix multiplications) and Non-MAC-Ops (other DNN operations) for TiM-DNN (TiM) and the baselines. Overall, we achieve 5.1x-7.7x speedup over the Iso-capacity baseline and 3.2x-4.2x speedup over the Iso-area baseline across our benchmark applications. The speedups depend on the fraction of application runtime spent on MAC-Ops, with DNNs having higher MAC-Ops times attaining superior speedups. This is to be expected, as the performance benefits of TiM-DNN over the baselines derive from accelerating MAC-Ops using in-memory computations. The Iso-area design is faster than the Iso-capacity design due to the higher-level of parallelism available from the additional baseline tiles. 

In absolute terms, the 32-tile instance of TiM-DNN achieves 4827, 952, 1834, 2*$10^{6}$, and 1.9*$10^{6}$ inference/sec for AlexNet, ResNet-34, Inception, LSTM, and GRU, respectively. The RNN benchmarks (LSTM and GRU) fit on TiM-DNN entirely, leading to better inference performance than CNNs. 

\begin{figure}[htb]
  \vspace*{-0pt}
  \centering
  \includegraphics[width=\columnwidth]{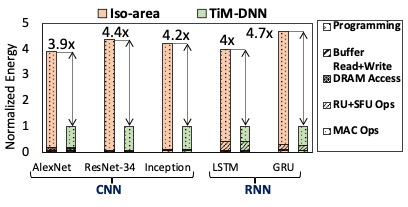}
  \vspace*{-6pt}
  \caption{Energy benefits of TiM-DNN}
  \label{fig:energyBenefit}
  \vspace*{-6pt}
\end{figure}

We next analyze the energy benefits of TiM-DNN over the superior of the two baselines (Iso-area design). Figure~\ref{fig:energyBenefit} shows the major energy components, which are programming (writes to TiM tiles), off-chip DRAM accesses, reads (writes) from (to) activation and Psum buffers, operations in reduce units and special function units (RU+SFU Ops), and MAC-Ops. As shown, TiM reduces the MAC-Ops energy substantially and achieves 3.9x-4.7x energy improvements across the benchmarks. The primary cause for this energy reduction is the in-memory computing capability of TiM-DNN.   

\subsection{Kernel-level benefits}
\label{subsec:KernelLevel}

\begin{figure}[htb]
  \vspace*{-0pt}
  \centering
  \includegraphics[width=\columnwidth]{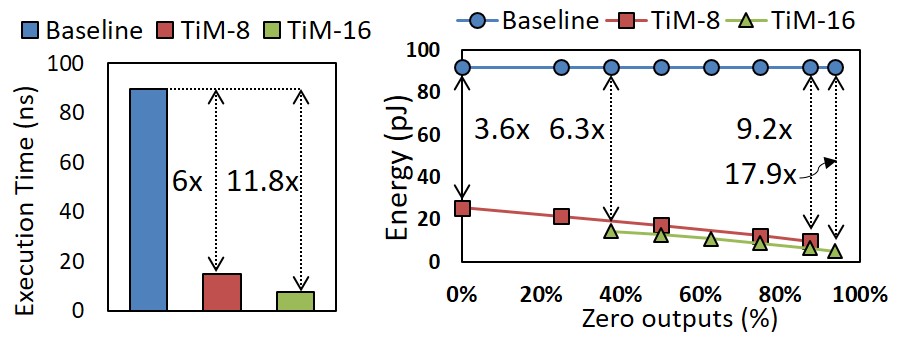}
  \vspace*{-6pt}
  \caption{Kernel-level benefits of TiM tiles}
  \label{fig:kernelLevel}
  \vspace*{-0pt}
\end{figure}

To provide further insights into the performance and energy benefits, we compare the TiM tile and the near-memory baseline tile at the kernel-level. We consider a primitive DNN kernel,~\emph{i.e.}, a vector-matrix computation (Out = Inp*W, where Inp is a 1x16 vector and W is a 16x256 matrix), and map it to both TiM and baseline tiles. We use two TiM tile variants, TiM-8 and TiM-16, wherein we simultaneously activate 8 wordlines and 16 wordlines, respectively. Using the baseline tile, the vector-matrix multiplication operation requires row-by-row sequential reads, resulting in 16 SRAM accesses. In contrast, TiM-16 and TiM-8 require 1 and 2 accesses, respectively. Figure~\ref{fig:kernelLevel} shows that the TiM-8 and TiM-16 designs achieve a speedup of 6x and 11.8x respectively, over the baseline design. Note that the benefits are lower than 8x and 16x, respectively, as single-row SRAM reads require lower access times than TiM-8 and TiM-16 accesses.

Next, we compare the energy consumption of TiM-8, TiM-16, and baseline tiles for the vector-matrix multiply computation. In TiM-8 and TiM-16, the bit-lines are discharged twice and once, respectively, whereas, in the baseline design the bit-lines discharge multiple (16*2) times. Therefore, TiM tiles achieve substantial energy benefits over the baseline design. The additional factor `2' in (16*2) arises as the SRAM array uses two 6T bit-cells for storing a ternary word. However, the energy benefits of TiM-8 and TiM-16 are not 16x and 32x, respectively, as TiM tiles discharge the bitlines by larger amounts (multiple $\Delta$s). Further, the amount by which the bitlines get discharged in TiM tiles depends on the number of non-zero scalar outputs. For example, in TiM-8, if 50\% of the TPCs output in a column are zeros the bitline discharges by $4\Delta$, whereas if 75\% are zeros the bitline discharges by only $2\Delta$. Thus, the energy benefits over the baseline design are a function of the output sparsity (fraction of outputs that are zero). Figure~\ref{fig:kernelLevel} shows the energy benefits of TiM-8 and TiM-16 designs over the baseline design at various output sparsity levels.

\subsection{TiM-DNN area breakdown}
\label{subsec:area}

\begin{figure}[htb]
  \vspace*{-6pt}
  \centering
  \includegraphics[width=\columnwidth]{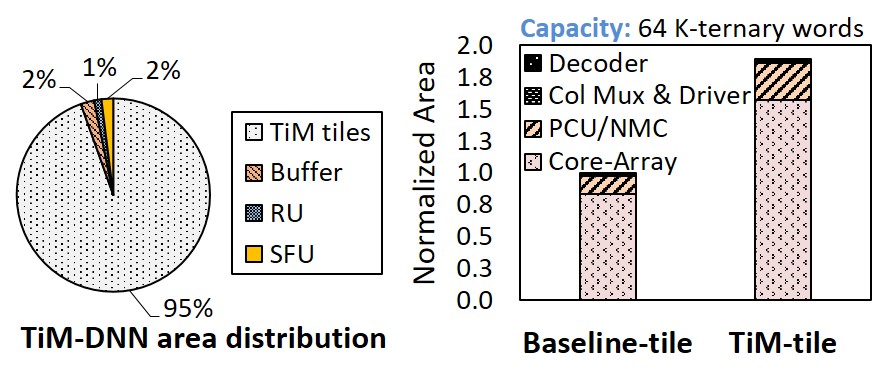}
  \vspace*{-6pt}
  \caption{TiM-DNN area breakdown}
  \label{fig:areaAnalysis}
  \vspace*{-0pt}
\end{figure}

We now discuss the area breakdown of various components in TiM-DNN. Figure~\ref{fig:areaAnalysis} shows the area breakdown of the TiM-DNN accelerator, a TiM tile, and a baseline tile. The major area consumer in TiM-DNN is the TiM-tile. In the TiM and baseline tiles, area mostly goes into the core array that consists of TPCs and 6T bit-cells, respectively. Further, as discussed in section~\ref{sec:exptsetup}, TiM tiles are 1.89x larger than the baseline tile at iso-capacity. Therefore, we use the iso-area baseline with 60 tiles and compare it with TiM-DNN having 32 TiM tiles.

\subsection{TiM-tile energy breakdown}
\label{subsec:area}

\begin{figure}[htb]
  \vspace*{-0pt}
  \centering
  \includegraphics[width=0.8\columnwidth]{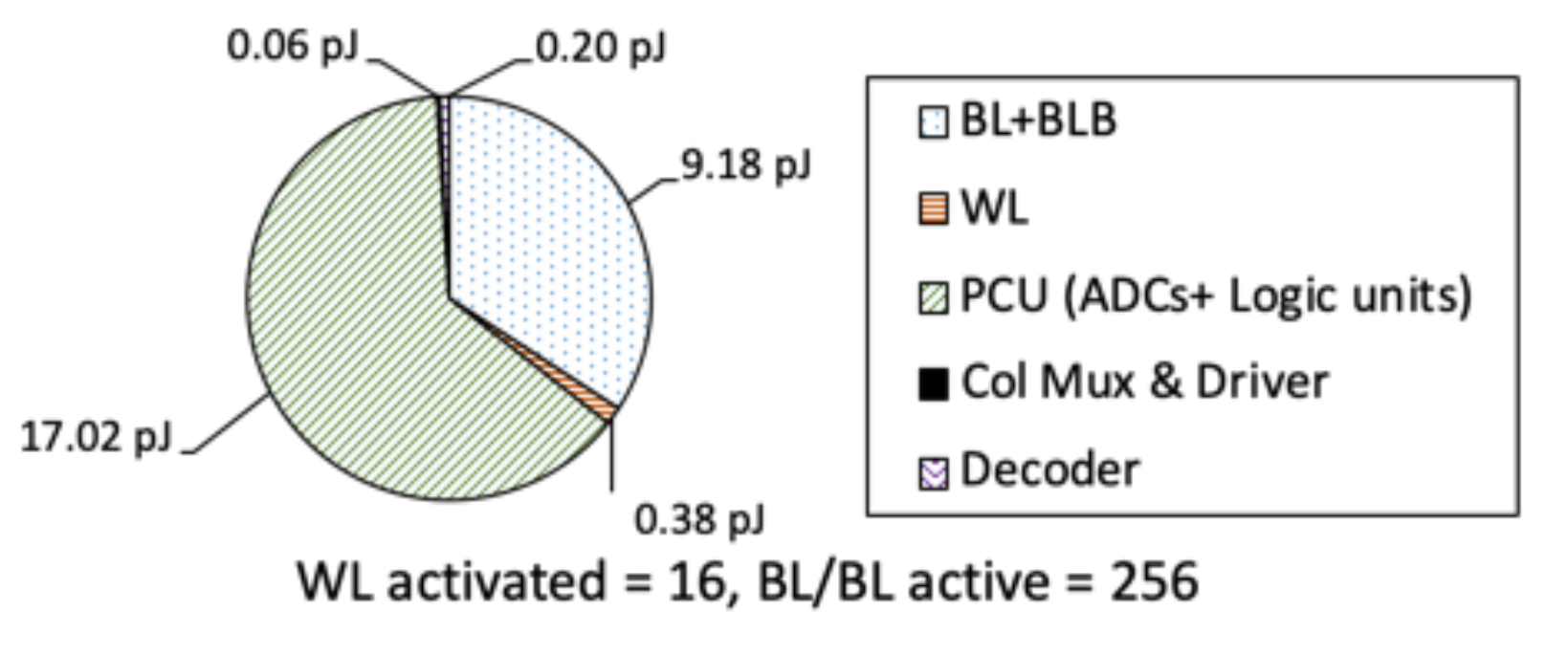}
  \vspace*{-6pt}
  \caption{Energy breakdown of a 16x256 ternary vector-matrix multiplication operation using TiM tile}
  \label{fig:energyBreakDown}
  \vspace*{-0pt}
\end{figure}

Next, we discuss the energy breakdown of various components in a TiM processing tile during a 16x256 ternary vector-matrix multiplication operation. Figure~\ref{fig:energyBreakDown} shows the major components of energy consumption,~\emph{viz.}, bitline energy (BL+BLB), wordline energy (WL), energy consumed in peripheral compute units that include ADCs (PCU), column mux and drivers, and decoders. Overall, a 16x256 ternary vector-matrix multiplication consumes 26.84pJ of energy. The most dominant component is the PCU (17pJ) due to 512 analog-to-digital conversion operations. The BL energy is also significant (9.18pJ). The WL energy component is quite small (0.38pJ), as the energy consumed by activating 16 word-lines pales in comparison to the energy consumed across 256 BLs and BLBs.

\begin{figure}[htb]
  \vspace*{-0pt}
  \centering
  \includegraphics[width=\columnwidth]{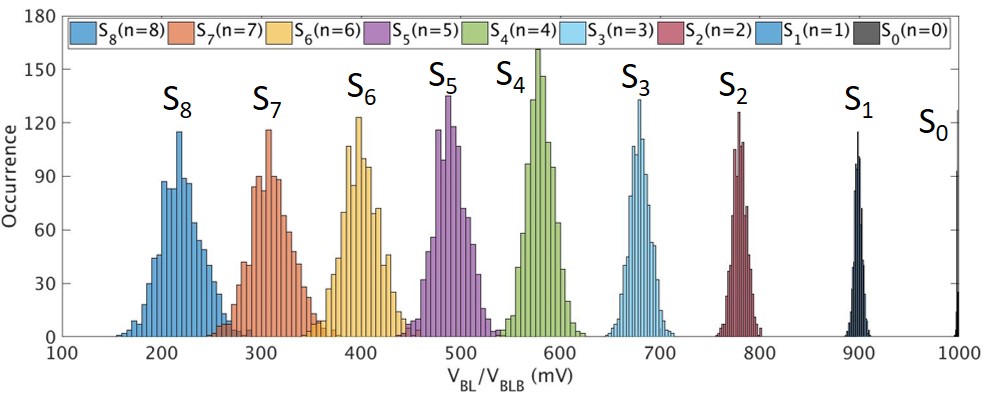}
  \caption{Histogram of the bit-line voltages ($V_{BL}$/$V_{BLB}$) under process variations}
  \vspace*{-6pt} 
 \label{tab:bitlineVolVariation}
  \vspace*{-0pt}
\end{figure}

\subsection{Impact of variations}
\label{subsec:varAnalysis}

Finally, we study the impact of process variations on the computations (~\emph{i.e.}, ternary vector-matrix multiplications) performed using TiM-DNN. To that end, we first perform Monte-Carlo circuit simulation of ternary dot-product operations executed in TiM tiles with $n_{max}$ = 8 and L = 16 to determine the sensing errors under random variations. We consider variations ($\sigma/\mu$ = 5\%)~\cite{kuhn2011} in the threshold voltage ($V_{T}$) of all transistors in the TPCs. We evaluate 1000 samples for every possible BL/BLB state ($S_{0}$ to $S_{8}$) and determine the spread in the final bitline voltages ($V_{BL}$/$V_{BLB}$). Figure~\ref{tab:bitlineVolVariation} shows the histogram of the obtained $V_{BL}$ voltages of all possible states across these random samples. As mentioned in section~\ref{subsec:dotProduct}, the state $S_{i}$ represents $n = i$, where $n$ is the ADC Output. We can observe in the figure that some of the neighboring histograms slightly overlap, while the others do not. For example, the histograms for $S_{7}$ and $S_{8}$ overlap but those for $S_{1}$ and $S_{2}$ do not. The overlapping areas in the figure represent the samples that will result in sensing errors. However, the overlapping areas are very small, indicating that the probability of the sensing error ($P_{SE}$) is extremely low. Further, the sensing errors depend on $n$, and we represent this dependency as the conditional sensing error probability [$P_{SE}(SE\mid n)$]. It is also worth mentioning that the error magnitude is always $\pm$1, as only the adjacent histograms overlap.

\begin{equation}
\begin{aligned}
\centering
   \begin{split}
 	&  P_{E} = \sum_{n=0}^{8}P_{SE}(SE\mid n)*P_{n}
	\end{split}
	\end{aligned}
	\label{eq:errorProb}
\end{equation}

\begin{figure}[htb]
  \vspace*{-0pt}
  \centering
  \includegraphics[width=\columnwidth]{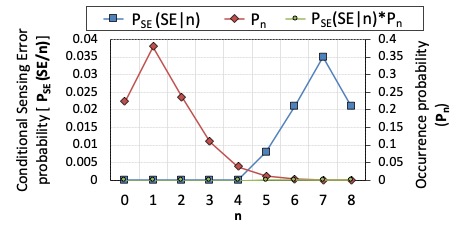}
  \caption{Error probability during vector-matrix multiplications}
  \vspace*{-6pt}
  \label{tab:errorProb}
  \vspace*{-0pt}
\end{figure}

Equation~\ref{eq:errorProb} details the probability ($P_{E}$) of error in the ternary vector-matrix multiplications executed using TiM tiles, where $P_{SE}(SE\mid n)$ and $P_{n}$ are the conditional sensing error probability and the occurrence probability of the state $S_{n}$ (ADC-Out = n), respectively.  Figure~\ref{tab:errorProb} shows the values of $P_{SE}(SE\mid n)$, $P_{n}$, and their product ($P_{SE}(SE\mid n)$*$P_{n}$) for each n. $P_{SE}(SE\mid n)$ is obtained using the Monte-Carlo simulation (described above), and $P_{n}$ is computed using the traces of the partial sums obtained from sample ternary DNNs~\cite{WRPN,HitNet}. As shown in Figure~\ref{tab:errorProb}, $P_{n}$ is maximum at $n=1$ and drastically decreases with higher values of $n$. In contrast, $P_{SE}(SE\mid n)$ shows an opposite trend, wherein the probability of sensing error is higher for larger $n$. Therefore, we find the product $P_{SE}(SE\mid n)$*$P_{n}$ to be quite small across all values on $n$. In our evaluation, $P_{E}$ is found to be 1.5*$10^{-4}$, reflecting an extremely low probability of error. In other words, we have roughly 2 errors of magnitude ($\pm$1) for every 10K ternary vector matrix multiplications executed using TiM-DNN. Through application-level simulations, we found that $P_{E}$ = 1.5*$10^{-4}$ has no impact on the DNN accuracy for our benchmarks. We note that this is due to the low probability and magnitude of error, as well as the ability of DNNs to tolerate errors in their computations~\cite{AxNN}.

\vspace*{-0pt}
\section{Conclusion}
\label{sec:conclusion}
{\noindent}Ternary DNNs are extremely promising due to their ability to achieve accuracy similar to full-precision networks on complex machine learning tasks, while enabling DNN inference at low energy. In this work, we present TiM-DNN, an in-memory accelerator for executing state-of-the-art ternary DNNs. TiM-DNN is a programmable accelerator designed using TiM tiles,~\emph{i.e.}, specialized memory arrays for realizing massively parallel signed vector-matrix multiplications with ternary values. TiM tiles are in turn composed using a new Ternary Processing Cell (TPC) that functions as both a ternary storage unit and a scalar multiplication unit. We evaluate an instance of TiM-DNN with 32 TiM tiles and demonstrate that it achieves significant energy and performance improvements over GPUs, current DNN accelerators, as well as a well-optimized near-memory ternary accelerator baseline.


\vspace*{-0pt}
\scriptsize
\bibliographystyle{unsrt}
\bibliography{references}
\begin{IEEEbiography}
[{\includegraphics[width=1.1in,height=1.25in,clip,keepaspectratio]{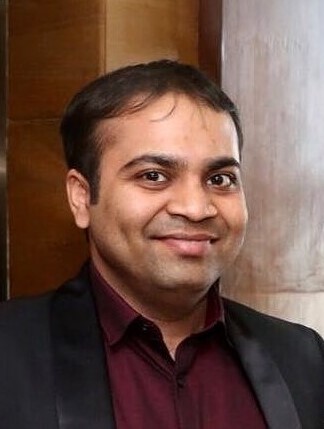}}]
{Shubham Jain} is currently a research staff member at IBM T.J. Watson Research Center, Yorktown Heights, New York. He has a B.Tech (Hons.) degree in Electronics and Electrical Communication Engineering from the Indian Institute of Technology, Kharagpur, India, in 2012, and a Ph.D. degree in Electrical and Computer Engineering from Purdue University, West Lafayette, Indiana, in 2019. His primary research interests include AI hardware, architecture for post-CMOS devices, in-memory computing, and approximate computing. Previously, he worked as a design engineer in the Bangalore Design Center, Qualcomm, Bangalore, India from 2012 to 2014. He also worked as a summer intern at IBM T.J Watson Research Center, Yorktown Heights, in 2017 and 2018. He has received the Mitacs Globalink scholarship from Mitacs, in 2011, the Andrews Fellowship from Purdue University, in 2014, and the A. Richard Newton Young Student Fellowship from DAC in 2015. His research has received the best technical paper award in DAC 2018, and a best-in session award in TECHCON 2016.
\end{IEEEbiography}

\begin{IEEEbiography}
[{\includegraphics[width=1.1in,height=1.25in,clip,keepaspectratio]{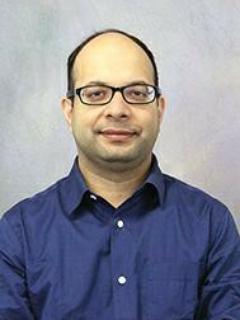}}]
{Sumeet Kumar Gupta} Sumeet Kumar Gupta received his Ph.D. degree from the School of Electrical and Computer Engineering, Purdue University, West Lafayette IN in 2012. He is currently an Assistant Professor of Electrical and Computer Engineering at Purdue University. Prior to this, he was an Assistant professor of Electrical Engineering at The Pennsylvania State University from 2014 to 2017 and a Senior Engineer at Qualcomm Inc. in San Diego CA from 2012 to 2014, where he developed circuit design techniques and methodologies for analysis and benchmarking of standard cells. His research interests include low power variation-aware VLSI circuit design, neuromorphic computing, memory design, and in-memory computing, nano-electronics and spintronics, device-circuit co-design and nano-scale device modeling. He has published over 90 articles in refereed journals and conferences. He was the recipient of DARPA Young Faculty Award in 2016, an Early Career Professorship by Penn State in 2014, the 6th TSMC Outstanding Student Research Bronze Award in 2012, Magoon Award and the Outstanding Teaching Assistant Award from Purdue University in 2007 and Intel Ph.D. Fellowship in 2009.
\end{IEEEbiography}
\vspace*{-25pt}

\begin{IEEEbiography}
[{\includegraphics[width=1.1in,height=1.25in,clip,keepaspectratio]{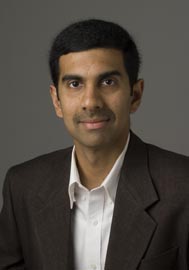}}]
{Anand Ragunathan} received the B. Tech. degree in Electrical and Electronics Engineering from the Indian Institute of Technology, Madras, India, and the M.A. and Ph.D. degrees in Electrical Engineering from Princeton University, Princeton, NJ. He is a Professor and Chair of the VLSI area in the School of Electrical and Computer Engineering at Purdue University, where he directs research in the Integrated Systems Laboratory and serves as the Associate Director of the SRC/DARPA Center for Brain-inspired Computing. His research interests include brain-inspired computing, energy-efficient machine learning and artificial intelligence, system-on-chip design and computing with post-CMOS devices. He also holds a Distinguished Visiting Chair in Computational Brain Research at the Indian Institute of Technology, Madras and co-founded High Performance Imaging, Inc. Before joining Purdue, he was a Senior Researcher and Project Leader at NEC Laboratories America and held a visiting position at Princeton University.

Anand has co-authored a book, eight book chapters, and over 250 refereed journal and conference papers, and holds 24 U.S patents and 16 international patents. His publications received nine best paper awards and six best paper nominations at premier IEEE and ACM conferences. He received a Patent of the Year Award and two Technology Commercialization Awards from NEC. He also received an IBM Faculty Award and Qualcomm Faculty Award. He was chosen by MIT's Technology Review among the TR35 (top 35 innovators under 35 years, across various disciplines of science and technology) in 2006. He also received the Distinguished Alumnus Award from IIT Madras. Prof. Raghunathan has chaired four premier IEEE/ACM conferences, and served on the editorial boards of various IEEE and ACM journals in his areas of interest. He received the IEEE Meritorious Service Award and Outstanding Service Award. He is a Fellow of the IEEE and Golden Core Member of the IEEE Computer Society. 
\end{IEEEbiography}

\end{document}